\documentclass[conference]{IEEEtran}
\usepackage[utf8]{inputenc}
\usepackage[english]{babel}
\usepackage{graphicx}
\usepackage{amsmath}
\usepackage{amssymb}
\usepackage{booktabs}
\usepackage{hyperref}
\usepackage{cite}
\usepackage{subfig}

\title{Few-Shot Learning of a Graph-Based Neural Network Model Without Backpropagation}

\author{\IEEEauthorblockN{Mykyta Lapin}
\IEEEauthorblockA{\textit{Department of System Analysis and} \\
\textit{Information and Analytical Technologies} \\
\textit{National Technical University "Kharkiv Polytechnic Institute"}\\
Kharkiv, Ukraine \\
Mykyta.Lapin@cit.khpi.edu.ua \\
ORCID: 0009-0003-6307-1172}
\and
\IEEEauthorblockN{Kostiantyn Bokhan}
\IEEEauthorblockA{\textit{Department of System Analysis and} \\
\textit{Information and Analytical Technologies} \\
\textit{National Technical University "Kharkiv Polytechnic Institute"}\\
Kharkiv, Ukraine \\
kostiantyn.bokhan@khpi.edu.ua \\
ORCID: 0000-0003-3375-2527}
\and
\IEEEauthorblockN{Yurii Parzhyn}
\IEEEauthorblockA{\textit{Postdoctoral Fellow} \\
\textit{School of Computer and Cyber Sciences} \\
\textit{Augusta University}\\
Augusta, USA \\
yparzhyn@augusta.edu \\
ORCID: 0000-0001-5727-1918}
}

\date{}

\begin{document}

\maketitle

\begin{abstract}
\textbf{Subject.} We propose a structural-graph approach to classifying contour images in a few-shot regime without using backpropagation. The core idea is to make structure the carrier of explanations: an image is encoded as an attributed graph (critical points and lines represented as nodes with geometric attributes), and generalization is achieved via the formation of concept attractors (class-level concept graphs). 
\textbf{Purpose.} To design and experimentally validate an architecture in which class concepts are formed from a handful of examples (5–6 per class) through structural and parametric reductions, providing transparent decisions and eliminating backpropagation. 
\textbf{Objectives.} (1) Define a vocabulary of node/edge types and an attribute set for contour graphs; (2) specify normalization and invariances; (3) develop structural and parametric reduction operators as monotonic structural simplifications; (4) describe a procedure for aggregating examples into stable concepts; (5) perform classification via graph edit distance (GED) with practical approximations; (6) compare with representative few-shot approaches. 
\textbf{Methods.} Contour vectorization is followed by constructing a bipartite graph (Point/Line as nodes) with normalized geometric attributes such as coordinates, length, angle, and direction; reductions include the elimination of unstable substructures or noise and the alignment of paths between critical points. Concepts are formed by iterative composition of samples, and classification is performed by selecting the best graph-to-concept match (using approximated GED). 
\textbf{Results.} On an MNIST subset with 5–6 base examples per class (single epoch), we obtain a consistent accuracy of around 82\% with full traceability of decisions: misclassifications can be explained by explicit structural similarities. An indicative comparison with SVM, MLP, CNN, as well as metric and meta-learning baselines, is provided. 
\textbf{Conclusions.} The structural-graph scheme with concept attractors enables few-shot learning without backpropagation and offers built-in explanations through the explicit graph structure. Limitations concern the computational cost of GED and the quality of skeletonization; promising directions include classification-algorithm optimization, work with static scenes, and associative recognition.
\end{abstract}


\section{Introduction}
Recent advances in Artificial Intelligence (AI), particularly in Deep Learning and Artificial Neural Networks (ANN), have led to significant progress in solving complex tasks [1--3]. However, the widespread application of these technologies has revealed a number of fundamental limitations that question the possibility of creating truly autonomous and adaptive systems [4--6]. These limitations include: the need for massive amounts of data for training, which requires significant time, computational, and energy resources [7, 8]; fundamental problems of generative models related to information trust, "hallucinations," and the "entropy gap" phenomenon [4, 7, 9]; and model degradation when training on recursively generated data (model autophagy disorder, MAD) [10, 11].

In this work, we proceed from the assumption that these problems are of a fundamental nature, stemming from the current conceptual paradigm. Modern ANNs are based primarily on the statistical nature of learning and a rigid architecture, which is optimized using the backpropagation algorithm [2, 3, 6]. Even specialized approaches for few-shot learning, such as meta-learning (MAML, Prototypical Networks) [12--14], are essentially complex methods of statistical optimization. They do not eliminate the fundamental dependence on statistics and cannot learn truly "from scratch" on a few examples, as they rely on models pre-trained on large data or require a complex meta-learning stage.

The paper considers an alternative approach based on abandoning backpropagation in favor of biologically grounded structural generalizations. This work presents a practical computational implementation of such an approach. We demonstrate how visual patterns (contour images) can be represented as attributed graphs [15--17], where nodes (critical points, lines) and edges (spatial connections) encode the topological and geometric properties of the object. The learning process is implemented as single-pass few-shot learning without backpropagation. It is based on the application of structural and parametric reduction operators, which act through monotonic structural simplification. Iterative application of these operators on 5-6 unique samples forces the system to converge to a stable, generalized state with minimal structural complexity — a generalized concept graph (or prototype graph).

\section{Analysis of Recent Research and Publications}
The development of structural-graph models for few-shot learning lies at the intersection of several key research directions: Few-shot learning [14], Explainable AI (XAI) methods [18, 19], Graph representations (GED) [20, 21], and alternative architectures (OvA/OvO) [22]. Analysis of the literature in these areas reveals fundamental conceptual limitations that the proposed approach aims to address [7, 23--25].

\subsection{Few-shot/Meta-learning}
Dominant deep learning models (CNN, MLP, Transformer) are fundamentally statistical and demonstrate low efficiency when training on critically small datasets, requiring thousands of examples and many training epochs to achieve acceptable accuracy. To address this problem, few-shot and meta-learning methods have been proposed [2, 7, 14, 26]. Prototypical Networks learn to identify class prototypes based on a distance metric in the embedding space [13]. MAML (Model-Agnostic Meta-Learning) attempts to find an optimal initial weight initialization for fast adaptation [12]. Although both methods significantly improve accuracy on small samples, they do not eliminate the fundamental dependence on statistics and backpropagation. They require a complex and resource-intensive meta-learning stage on large auxiliary datasets [14, 26]. Thus, this is a transfer of knowledge obtained statistically, rather than true "from scratch" single-pass learning.

\subsection{Explainable AI (XAI) and Graph Representations}
Simultaneously with the increasing complexity of models, the problem of their interpretability has intensified. Deep learning models function as "black boxes". Popular XAI methods such as LIME and SHAP are post-hoc techniques: they attempt to approximate the behavior of an already trained model rather than explain its actual decision-making process [27, 28]. Studies have shown that such explanations can be unreliable, contradictory, and vulnerable to adversarial attacks [18, 19, 29, 30]. An alternative is "explainability by design," where the internal representation of the model is semantically meaningful [16, 18, 19]. Graph structures are an ideal candidate for this, as they allow explicit encoding of semantics in nodes and edges. Graph Edit Distance (GED) [20, 21] is used to compare such structures. However, GED is an NP-hard problem, which remains a challenge for practical application [31, 32].

\subsection{Alternative Architectures (OvA/OvO) and Feature Generalization Problems}
Alternative ANN architectures -- "One-vs-All" (OvA) and "One-vs-One" (OvO) -- have long been considered for classification tasks [22]. This is an approach where instead of one large network, specialized networks are used (e.g., one for each class). This approach is conceptually close to ours, where we build one separate "neuron" (concept graph) for each class. However, in classical implementations of OvA/OvO architectures relying on backpropagation, noticeable limitations are observed regarding Out-of-Distribution Detection (OOD) [33, 34]. Networks trained on limited examples do not form stable class separation boundaries. This is because traditional ANNs generalize only local recognition features (e.g., individual textures or angles) and cannot generalize features at the level of the entire structure [23, 24]. Their fully connected and combinatorial nature with stochastic initialization makes generalization of global, topological properties impossible. Our approach solves this problem because generalization occurs not through stochastic optimization of local weights, but through deterministic structural graph reduction, which captures global topological features.

\subsection{Synthesis: Identified Conceptual Gaps}
Literature analysis reveals three distinct but interconnected problems:
\begin{enumerate}
    \item Dependence of few-shot learning methods on backpropagation: Leading few-shot methods (MAML, ProtoNets) are not true "learning from scratch" but knowledge transfer methods requiring intensive pre-training using backpropagation.
    \item Unreliability of XAI: Existing XAI methods (LIME, SHAP) remain largely post-hoc, unreliable, and vulnerable to attacks.
    \item Locality of features in OvA: Classical architectures (including OvA) capture local patterns but are unable to generalize global/structural features, leading to OOD problems and unstable decision boundaries.
\end{enumerate}

\section{Unresolved Issues, Purpose, and Objectives}
Unresolved Issues (research gaps):
\begin{enumerate}
    \item Reliability of explanations: Approaches with "explainability by design" are needed, rather than post-hoc approximations (LIME/SHAP).
    \item Few-shot learning without backpropagation: Leading methods (MAML, ProtoNets) still rely on gradient updates. Alternatives working in low-data regimes without backprop are needed.
    \item Generalization of global features: Classical ANNs (including OvA) capture local patterns but fail to generalize global topological structure, which is key to shape recognition.
    \item Complexity of GED: Structural graph matching (GED) is NP-hard, limiting its practical application.
\end{enumerate}

\textbf{Purpose of the Work} -- To develop and experimentally validate a structural-graph approach to few-shot classification of contour images without backpropagation, in which the generalization of several class examples is performed through a sequence of structural and parametric reductions, and decision-making has built-in explainability thanks to the explicit graph structure.

\textbf{Objectives:}
\begin{enumerate}
    \item \textbf{Representation.} Define the representation of a contour image as an attributed graph (node/edge types, geometric attributes, normalization, and invariances) taking into account skeletonization/vectorization properties.
    \item \textbf{Reduction Operators.} Develop a set of structural (removing unstable branches, merging intersections, normalizing paths) and parametric (min--max--center ranges for numerical features) operators that simplify a set of examples into a concept attractor.
    \item \textbf{Example Aggregation.} Build a procedure for forming a concept from 5--6 examples per class in few-shot mode, fixing attribute tolerances and filtering random structures.
    \item \textbf{Classification.} Design a matching scheme with concepts (GED with heuristics based on bipartite matching/local searches) with hard time and quality constraints.
    \item \textbf{Experimental Protocol.} Conduct tests on a subset of MNIST/similar contour sets: one epoch, 5--6 base examples/class (+augmentations); evaluate accuracy, concept stability, computation time.
    \item \textbf{Comparison with FSL Baselines.} Compare with representative methods (Prototypical Networks, MAML) as examples of metric and meta-learning approaches; present an indicative chart (caveats regarding different protocols).
    \item \textbf{Explainability and Risks.} Explicitly record structural subgraphs/attributes supporting the decision; compare with post-hoc explanations and discuss validity limits (when structure "does not explain").
\end{enumerate}

\section{Materials and Methods}
This section details the methodological pipeline used to convert 2D contour images into stable concept graphs and their subsequent classification. The methodology is based on principles of structural generalization and abandons gradient optimization.

\subsection{Representation of Contours as Attributed Graphs}
To achieve transparency and move away from "opaque" weight matrices inherent in traditional neural networks, a representation is proposed where "structure is the carrier of explanations." The input contour image, obtained after binarization and skeletonization stages, is transformed into an attributed graph. The system encodes contours as bipartite graphs, whose structure strictly alternates between nodes of type \textbf{Point} and nodes of type \textbf{Line}. This architectural differentiation is fundamental as it allows clear separation of topological structure (critical points) from geometric properties (segments connecting them).

Point Nodes: Represent topological structure and critical contour points. They are ontologically classified into four main types:
\begin{itemize}
    \item \textbf{EndPoint:} Terminal nodes marking the start or end of an open contour.
    \item \textbf{CornerPoint:} Nodes marking sharp changes in direction (corners).
    \item \textbf{IntersectionPoint:} Nodes where three or more segments connect.
    \item \textbf{StartPoint:} A dedicated anchor node defining the canonical starting point of graph traversal to ensure comparison consistency.
\end{itemize}

\textbf{Line Nodes:} Represent geometric properties. Importantly, line segments are represented as first-class nodes, not edges. This allows assigning them rich semantic and geometric attributes on par with Point nodes, which is critical for subsequent parametric reduction operations.

Edges (Relationships): Point and Line nodes are connected exclusively by bidirectional CONNECTED\_TO edges. This creates a strict traversal pattern Point $\rightarrow$ Line $\rightarrow$ Point $\rightarrow \dots$.

Each node carries a set of attributes encoding measurable geometry and topology parameters, specifically: \texttt{normalized\_x}, \texttt{normalized\_y} (coordinates normalized to invariant range [-1, 1]), \texttt{length} (segment length), \texttt{angle} (for CornerPoint), \texttt{quadrant} (discretized direction), \texttt{horizontal\_direction}, and \texttt{vertical\_direction}.

\subsection{Invariance through Normalization}
To ensure representation invariance to scale and shift, a necessary condition for stable attractor formation, all coordinates and related metrics (e.g., \texttt{length}) undergo normalization. Point coordinates are transformed into a centered system with range [-1, 1] using the formula:
\[
x_{normalized} = \frac{x - center\_x}{center\_x}
\]
A similar formula applies to $y$. This process is the first step of parametric reduction ($R_{uc}$), translating absolute, instance-specific values into relative, generalized parameters.

\subsection{Learning Process as Structural Graph Reduction}
The learning process (concept formation) in this work fundamentally differs from traditional statistical optimization (e.g., gradient descent on a loss function). It is viewed as a deterministic process of structural generalization striving towards a state of minimal structural complexity. This most stable, generalized system state representing the invariant essence of a class (e.g., all variants of writing the digit "3") is called the \textbf{generalized concept graph}.

The transition from a set of individual sample graphs ($G_1, \dots, G_n$) to a single concept graph $C$ is a process of controlled structure simplification (reduction). This process is governed by a set of \textbf{Custom Reduction Operations (CRO)}, which act by reducing structural complexity or parametric variability, attempting to simplify the graph to a stable prototype in a finite number of steps.

The general reduction process can be described as a composition of three classes of operators: 
\[
R(G_{input}) = R_w(R_{sp}(R_{u,c}(G_{input})))
\]
where $G_{input}$ is the initial graph, and $R_{uc}$, $R_{sp}$, $R_w$ are theoretical reduction operators. A key aspect of our methodology is the direct mapping of these theoretical operators to specific CRO algorithms implemented in the system, as detailed in Table 1.

\begin{table}[h]
\centering
\caption{Structural and Parametric Reduction Operators (CRO)}
\begin{tabular}{|p{0.25\linewidth}|p{0.25\linewidth}|p{0.4\linewidth}|}
\hline
\textbf{Theoretical Operator} & \textbf{Name and Goal} & \textbf{Practical Implementation (CRO) Details} \\
\hline
$R_{u,c}$ (Parametric Reduction) & \textbf{Parametric Generalization.} Minimization of parametric variability. Transition from quantitative values to generalized qualitative ranges. & \textbf{Numeric properties:} $v_1, \dots, v_n$ merge into range $s = \{min: v_{min}, max: v_{max}, center: v_{avg}\}$. Generalizes variations (e.g., length, angle). \newline \textbf{Categorical properties:} $L_1, \dots, L_n$ merge into $L$ only if $L_i = L_j$ for all $i, j$. Otherwise, attribute is removed. \newline \textbf{List properties:} $L_1, \dots, L_n$ merge via intersection $\cap$. \\
\hline
$R_{sp}$ (Structural-Parametric Reduction) & \textbf{Path Pruning.} Simplification (reduction) of structure based on the stability of its parameters. & For two aligned critical points, the algorithm finds all simple paths between them. It selects the "best match" based on node similarity and uses the shorter path as a template. Nodes from the longer path without a match are removed. This "eliminates length variations". \\
\hline
$R_{w}$ (Structural Reduction) & \textbf{Endpoint Removal and Intersection Point Merging.} Removal of topological elements that are statistically insignificant (noise). & \textbf{Endpoint removal:} Calculates similarity matrix of endpoints between concept $C_i$ and sample $G_{i+1}$. Endpoints with low similarity (below threshold) or "redundant" points are removed along with the entire path to the nearest critical point. \newline \textbf{Intersection merging:} Consolidates IntersectionPoint nodes representing the same structural feature. Applies semantic reduction (e.g., IntersectionPoint with degree $< 2$ becomes CornerPoint or EndPoint). \\
\hline
\end{tabular}
\end{table}

\subsection{Iterative Attractor Formation Algorithm}
The learning process is one-pass and does not require backpropagation. It iteratively builds an attractor based on an ultra-small sample consisting of 5-6 unique training samples per class. The concept is initialized with the first sample graph: $C = G_1$. This sample acts as an initial hypothesis about the class structure. Each subsequent sample $G_{i+1}$ is integrated into the current concept $C_i$ using the reduction operation: 
\[
C_{i+1} = CRO(C_i + G_{i+1})
\]
Each $CRO$ operation is a five-stage process applying reduction operators from Table 1:
\begin{enumerate}
    \item Start Point Alignment: Establishing a common origin for traversing graphs $C_i$ and $G_{i+1}$ by clustering and selecting StartPoint.
    \item Critical Point Pre-processing: Applying structural operators $R_w$ (Endpoint removal, Intersection merging) to achieve basic structural compatibility.
    \item Traversal Synchronization: Generating synchronized paths between corresponding critical points in both graphs.
    \item Common Structure Identification: Applying the structural-parametric operator $R_{sp}$ (Path pruning) to normalize paths and eliminate length variations between critical points.
    \item Parametric Merging: Applying the parametric operator $R_{uc}$ (Parametric Generalization) to merge attributes of nodes surviving structural reduction.
\end{enumerate}

This iterative process is path-dependent; the order of sample presentation affects the final concept graph. This mimics a process where an initial hypothesis ($C_0$) is iteratively refined under the influence of new data ($G_{i+1}$), which acts as a reduction force, eliminating sample-specific variations (noise) and leaving only the generalized core.

\subsection{Classification via Approximated Graph Matching (GED)}
The classification (inference) process consists of comparing a graph $G_{test}$, obtained from an unknown input image, with each concept graph $C_k$ from the trained library, minimizing the Graph Edit Distance (GED) to the input graph $G_{test}$: 
\[
Class(G_{test}) = \arg\min_k GED(G_{test}, C_k)
\]
GED is defined as the minimum cost of a sequence of operations (insertion, deletion, substitution of nodes/edges) required to transform $G_{test}$ into $C_k$. To ensure GED correctly accounts for the generalized nature of concepts, we use custom cost functions.

\textbf{Node Substitution Cost:} The cost of replacing node $u \in G_{test}$ with node $v \in C_k$ is calculated based on "range-based cost functions":
\begin{itemize}
    \item For numerical attributes (e.g., \texttt{length}, \texttt{angle}): If the attribute value of $u$ (e.g., \texttt{u.length}) falls within the learned range of attribute $v$ (e.g., \texttt{v.length} \{min, max\}), the substitution cost for this attribute is 0. If the value is outside the range, the cost is proportional to the distance to the nearest range boundary.
    \item For categorical attributes: Cost is 0 for exact match or infinite (high) for mismatch.
    \item Label Compatibility: Substitution cost is infinite if base node types are incompatible (e.g., Line to Point).
\end{itemize}

\textbf{Edge Edit Cost:} Reduced cost to prioritize topological differences (presence/absence of nodes) over connectivity differences.

Calculating exact GED is an NP-hard problem. To ensure practical applicability, an approximation is used via a strict 60-second timeout for each individual comparison $GED(G_{test}, C_k)$. This timeout acts as a heuristic approximation, interrupting the search for the optimal edit path if it takes too long, and returning the best distance found so far.

\subsection{Classification Mechanism and Winner Selection}
The proposed architecture implements an approach conceptually close to One-vs-All, where each class $k$ is represented by a separate "neuron" which is the generalized concept graph $C_k$. The classification (inference) process consists of comparing the contour graph $G_{test}$ with each concept graph $C_k$ from the trained library. Unlike stochastic networks where neuron "excitation" is a numerical output (e.g., softmax), in our system, the "excitation" of the k-th neuron is the process of calculating the edit distance $GED(G_{test}, C_k)$. To select the final classification result, we apply the Winner-Takes-All concept. The winner is the class (concept) $C_k$ whose edit distance to the input graph ($G_{test}$) is minimal.
\[
Class(G_{test}) = \arg\min_{k} \{GED(G_{test}, C_1), \dots, GED(G_{test}, C_k)\}
\]
If the distance is equal for multiple classes, a conflict resolution rule applies. The class that is structurally more complex is chosen. Complexity is calculated as the sum of nodes and edges of the graph (Fig. 1).

\begin{figure*}[h]
    \centering
    \includegraphics[width=\textwidth]{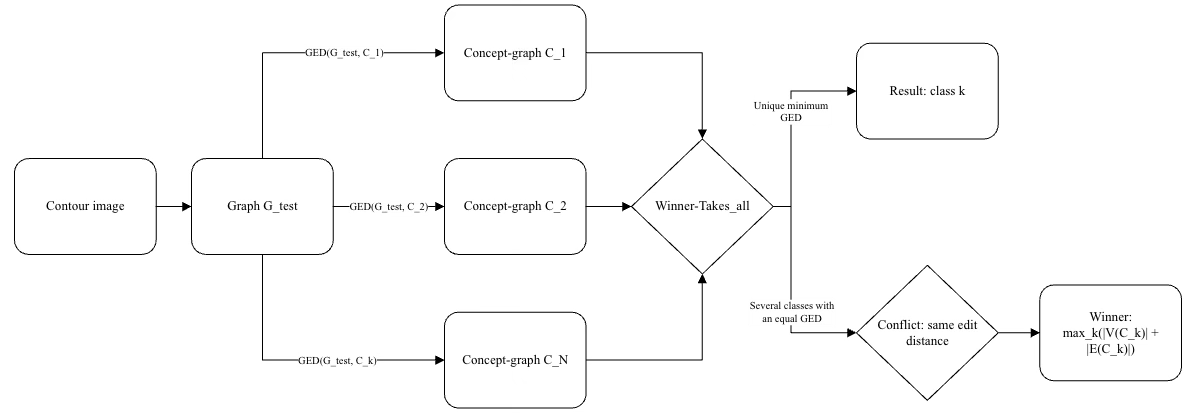}
    \caption{Classification scheme using GED and WTA.}
    \label{fig:classification}
\end{figure*}

\section{Results and Discussion}
This section presents empirical validation of the proposed graph approach to concept formation. The goal is not to optimize absolute accuracy, but to demonstrate that stable, explainable concept attractors can be formed from extremely limited data (few-shot learning) and that their performance and error patterns stem directly from their topological and parametric structure.
Experiments are conducted on the MNIST-6 subset (classes '1', '2', '3', '6', '7', '9'), using 5-6 unique training samples per subclass.

\subsection{Classification Efficiency on MNIST-6 in Few-Shot Regime}
The system was trained on 8 concepts covering 6 classes (some classes, like '1' and '2', had two concepts to represent different writing styles). Training consisted of iterative structural reduction of 5-6 base samples (with 10 augmentation variants per sample, totaling about 350 examples) for each concept. Evaluation was performed on a test set of 5467 images not involved in concept formation. General performance metrics are presented in Table 2.

\begin{table}[h]
\centering
\caption{Overall Classification Performance (5467 test images)}
\begin{tabular}{lc}
\toprule
\textbf{Metric} & \textbf{Value (\%)} \\
\midrule
Accuracy & 82.35 \\
Precision & 83.28 \\
Recall & 82.35 \\
F1 Score & 82.16 \\
\bottomrule
\end{tabular}
\end{table}

These results are conceptually significant. An accuracy of 82.35\% demonstrates that the approach based on forming canonical structural attractors without gradient optimization is viable and provides meaningful classification. The processing pipeline showed high reliability, with a 100\% success rate, except for 10 images (0.18\%) which failed processing due to skeletonization errors resulting in disconnected graphs.

\subsection{Per-Class Performance and Topological Distinctiveness Analysis}
In-depth analysis of metrics for each class (Table 3) reveals a direct dependence of performance on the structural uniqueness of digits.

\begin{table}[h]
\centering
\caption{Per-Class Classification Metrics}
\begin{tabular}{ccccc}
\toprule
\textbf{Digit} & \textbf{Precision (\%)} & \textbf{Recall (\%)} & \textbf{F1 (\%)} & \textbf{Count} \\
\midrule
1 & 81.46 & 96.49 & 88.34 & 997 \\
2 & 84.17 & 60.02 & 70.07 & 948 \\
3 & 78.21 & 87.28 & 82.50 & 983 \\
6 & 94.23 & 78.09 & 85.40 & 753 \\
7 & 74.38 & 82.12 & 78.06 & 990 \\
9 & 91.55 & 89.57 & 90.55 & 786 \\
\bottomrule
\end{tabular}
\end{table}

\textbf{Key observations:}
\begin{enumerate}
    \item High Precision for '6' (94.23\%) and '9' (91.55\%): These classes have the most unique topological signatures — closed loops represented by IntersectionPoint nodes. Their attractors are very specific, minimizing false positives.
    \item Low Recall for '2' (60.02\%): This indicator suggests that a significant portion (almost 40\%) of true '2' digits was not recognized. This indicates high morphological variability in writing '2', which the formed concepts ('2\_1' and '2\_2') could not fully capture. Their parametric ranges, learned from only 5-6 samples, proved too rigid.
    \item Low Precision for '7' (74.38\%): This class was most often confused with others, indicating its structural ambiguity, especially relative to digit '1'.
\end{enumerate}

\subsection{Confusion Matrix Analysis}
The confusion matrix (Figure 2) provides deep insight into how the model makes decisions, visualizing systematic errors that are a direct consequence of structural and topological similarity.

\begin{figure}[h]
    \centering
    \includegraphics[width=0.8\linewidth]{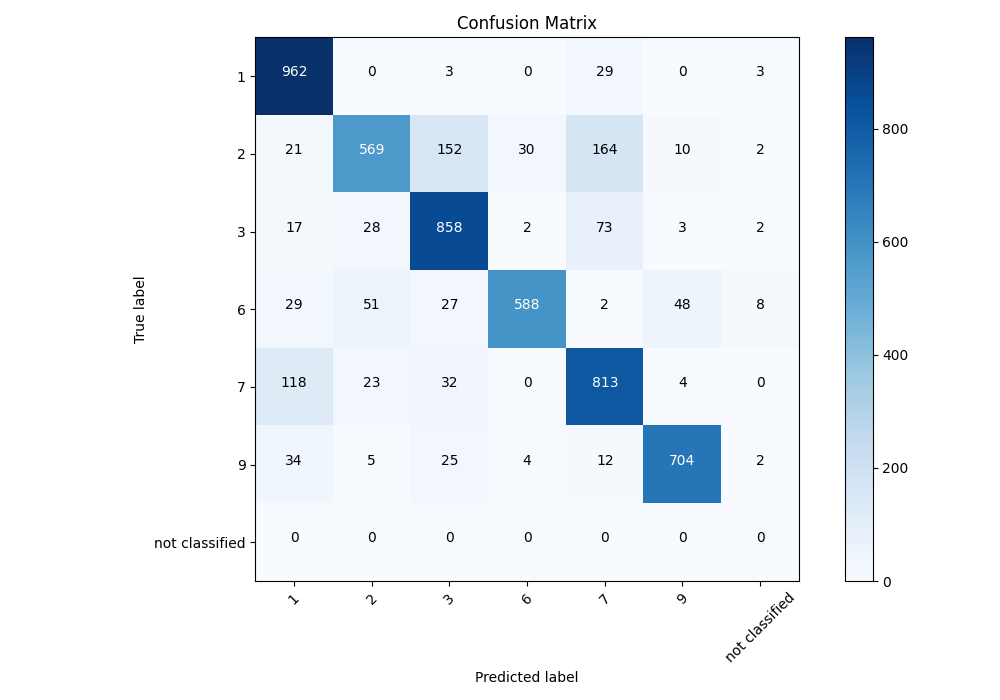}
    \caption{Confusion Matrix for 6-class MNIST classification.}
    \label{fig:confusion}
\end{figure}

Primary confusion occurs between digits 7 and 1 (angular open contours), and secondary confusion between digits 2 and 3 (curved open contours). Digits with a closed contour (6, 9) demonstrate strong discrimination.
\begin{itemize}
    \item Main confusion: 152 samples of digit '2' were classified as '3'. 28 samples of '3' were classified as '2'.
    \item Secondary confusion: 118 samples of digit '7' were classified as '1'.
\end{itemize}

Classes '6' and '9' demonstrate minimal confusion between themselves and with other open contours (e.g., only 48 samples of '6' were erroneously classified as '9').
Unlike "black boxes," where error causes are hidden in millions of weights, errors in this model are fully interpretable. Analysis shows that errors concentrate along structurally similar pairs:
\begin{enumerate}
    \item \textbf{'2' vs '3':} Both digits have similar "curved morphology." They are open contours starting from one side, having several bends (represented by CornerPoint nodes) and ending on the other side.
    \item \textbf{'7' vs '1':} Both digits are "angular open contours." They are both simple paths consisting of StartPoint, CornerPoint, and EndPoint. Confusion arises when writing '7' is less curved, or '1' has a more pronounced angle at the beginning.
\end{enumerate}
The fact that the model confuses '7' with '1' (structurally similar), but not '7' with '6' (structurally different — open vs closed contour), is powerful evidence that the graph matching mechanism works correctly and makes decisions based on topology, as designed.

\subsection{Concept Attractor Stability and Structural Explainability (XAI)}
This subsection analyzes the final result of the learning process — stable concept attractors, which are the carriers of explanations in the system. The structural reduction process transforms multiple training graphs into single canonical structures. Their metrics (Table 4) quantitatively define the "ideal" shape of each digit.

\begin{table}[h]
\centering
\caption{Structural Metrics of Concept Attractors (EP=EndPoint, CP=CornerPoint, IP=IntersectionPoint, SP=StartPoint)}
\begin{tabular}{ccccl}
\toprule
\textbf{Concept} & \textbf{Nodes} & \textbf{Edges} & \textbf{Avg. Degree} & \textbf{Critical Points} \\
\midrule
1\_1 & 3 & 2 & 1.33 & 1 EP, 1 SP \\
1\_3 & 3 & 2 & 1.33 & 1 EP, 1 SP \\
2\_1 & 7 & 6 & 1.71 & 1 EP, 2 CP, 1 SP \\
2\_2 & 12 & 12 & 2.00 & 1 EP, 3 CP, 1 IP, 1 SP \\
3\_1 & 7 & 6 & 1.71 & 1 EP, 2 CP, 1 SP \\
6\_1 & 10 & 10 & 2.00 & 3 CP, 1 IP, 1 SP \\
7\_1 & 5 & 4 & 1.60 & 1 EP, 1 CP, 1 SP \\
9\_2 & 8 & 8 & 2.00 & 2 CP, 1 IP, 1 SP \\
\bottomrule
\end{tabular}
\end{table}

Analysis of Table 4 demonstrates a direct correlation between digit topology and attractor complexity.
\begin{itemize}
    \item Concepts '1\_1' and '1\_3' are minimal, consisting of only 3 nodes (StartPoint, Line, EndPoint). This ideally reflects their topology as a simple, unbranched path.
    \item Concepts '6\_1' and '9\_2' have a higher average degree (2.00), indicating the presence of cycles. Importantly, they contain no EndPoint (EP=0), but contain IntersectionPoint (IP=1) where the cycle closes.
    \item Concepts '2\_1', '2\_2', '3\_1', '7\_1' have intermediate complexity (5-12 nodes). All contain exactly one EndPoint (EP=1), topologically marking them as open contours. The number of CornerPoints (CP) encodes the number of bends (e.g., '7\_1' has 1 CP, '2\_1' has 2 CP).
\end{itemize}

This table is essentially a dictionary for XAI. The explanation for classifying '9' is that the input image graph successfully matched concept '9\_2', which is canonically defined as an 8-node structure with 1 IntersectionPoint (cycle) and 0 EndPoints (no free ends).

\subsection{Case Study: Iterative Attractor Stabilization (Digit '3')}
The concept formation process (Figures 3a-d) is an empirical demonstration of theoretical reduction operators.
\begin{figure}[h]
    \centering
    \subfloat[]{\includegraphics[width=0.45\linewidth]{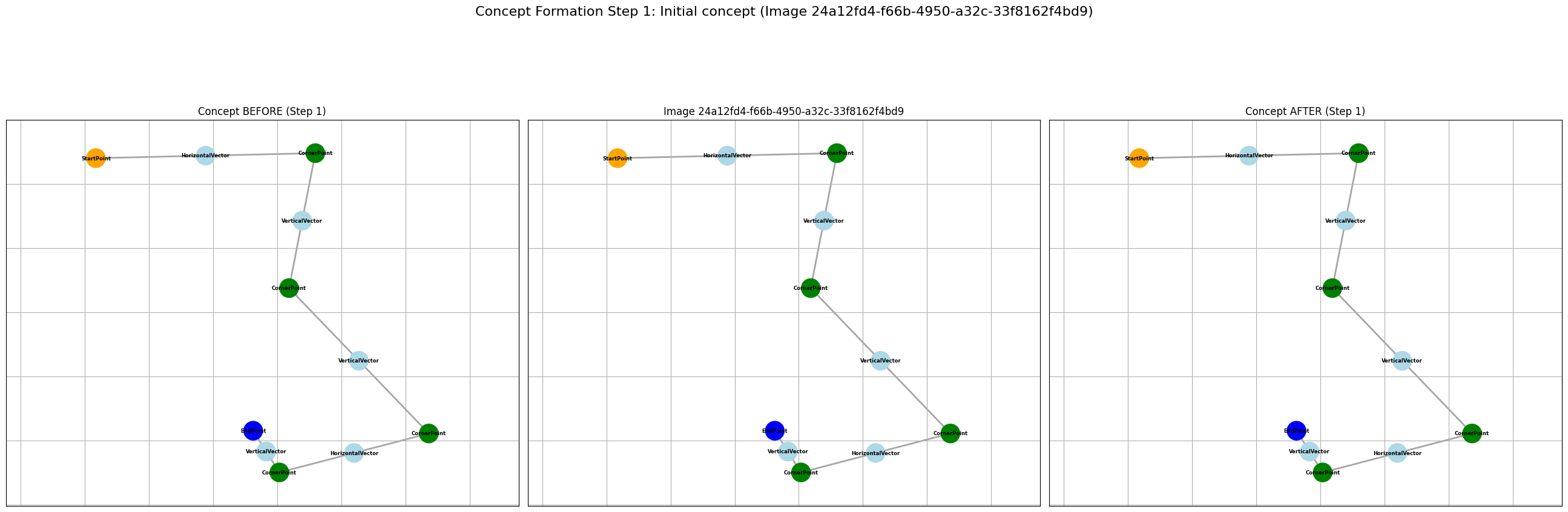}\label{fig:3a}}
    \hfill
    \subfloat[]{\includegraphics[width=0.45\linewidth]{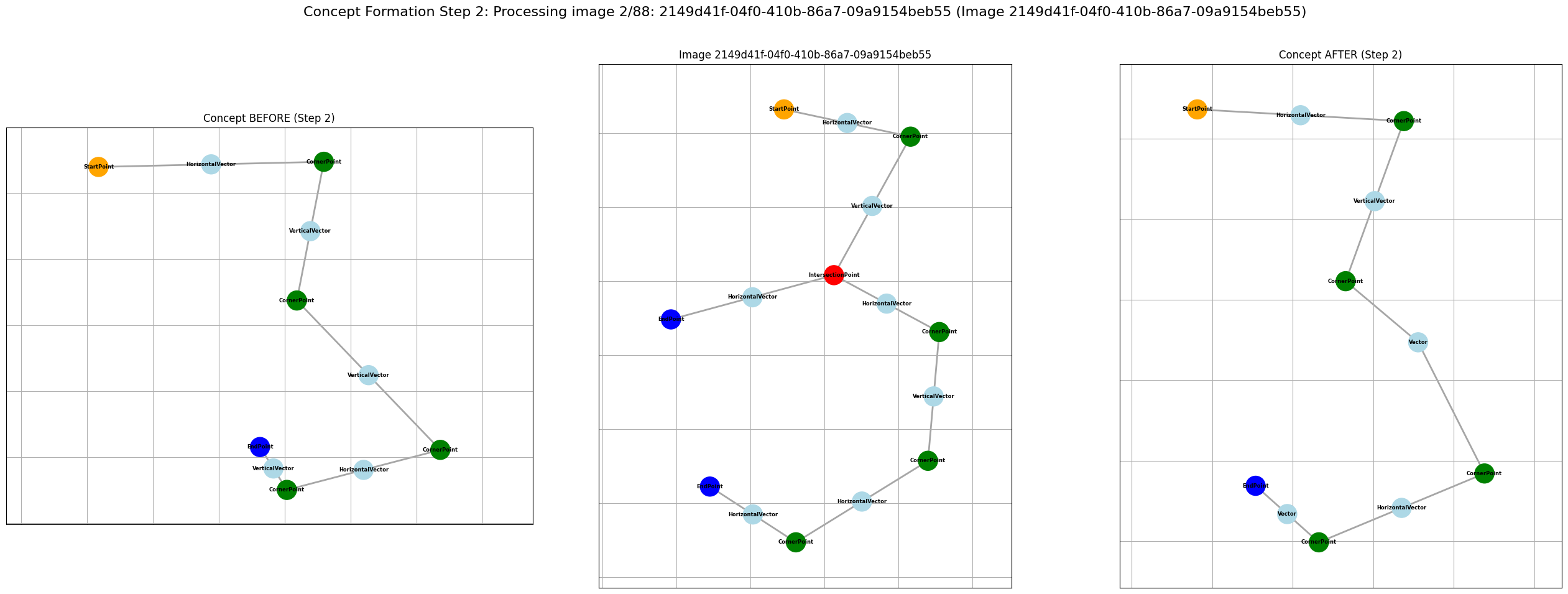}\label{fig:3b}}
    \\
    \subfloat[]{\includegraphics[width=0.45\linewidth]{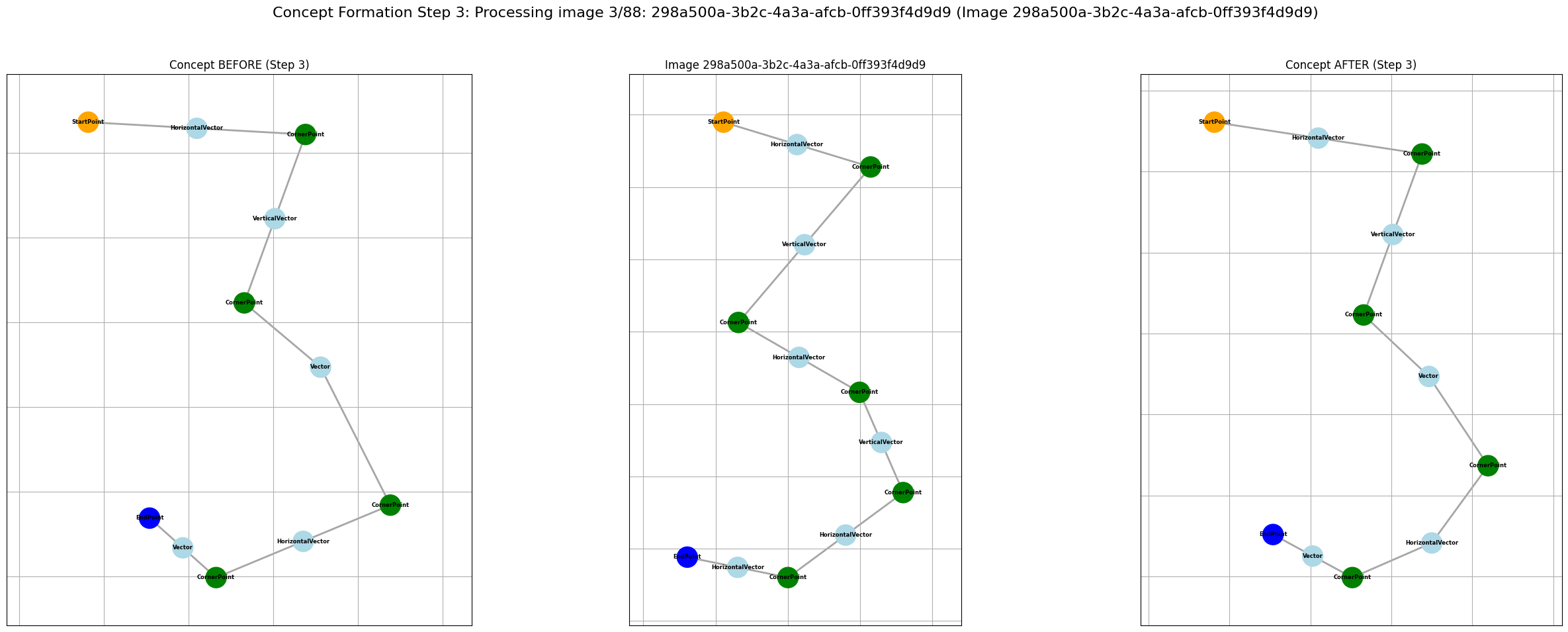}\label{fig:3c}}
    \hfill
    \subfloat[]{\includegraphics[width=0.45\linewidth]{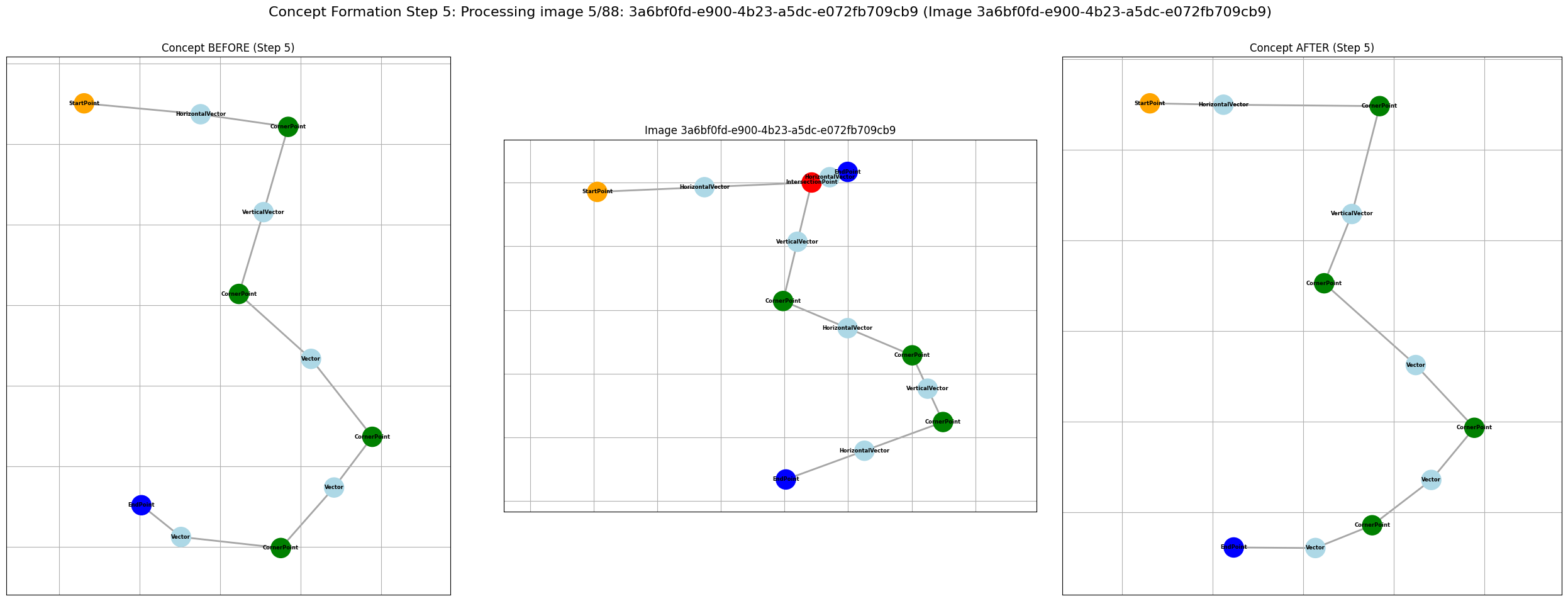}\label{fig:3d}}
    \caption{Concept formation process.}
    \label{fig:concept_formation}
\end{figure}

\textbf{Step 1 ($C_0 = G_1$):} The first sample ($G_1$) establishes the initial concept $C_0$. It is overly specific and contains all structural details and noise of the initial sample (Fig. 3a).
\textbf{Step 2 ($C_1 = CRO(C_0 + G_2)$):} Integrating the second sample ($G_2$) reveals a mismatch — a "redundant endpoint branch." The structural reduction operator (Endpoint removal) is applied, removing this $G_1$-specific noise. This is a practical implementation of the $R_w$ operator finding a common substructure (Fig. 3b).
\textbf{Step 3 and 4 ($C_2, C_3$):} Subsequent iterations continue this process, removing a "redundant corner point" (Fig. 3c) and other "noise substructure" (Fig. 3d).
The final concept $C_3$ (Fig. 3d) is a stable attractor representing the most general topological structure ("curved S-shape") common to all training samples. This process is a form of learning without backpropagation, where the representation structure itself is optimized, not a weight vector.

\subsection{Example of Parametric Generalization (Digit '3')}
Structural reduction determines which nodes remain, while parametric generalization determines how their attributes are generalized to encode variability. Using concept '3\_1' (formed from 3 samples) as an example:

Numerical Properties: Attributes like coordinates are not averaged but converted into ranges (\{min, max, center\}). This creates flexible decision boundaries.
\begin{itemize}
    \item $x_{normalized}$: [-0.7, 0.2] (center -0.33)
    \item $y_{normalized}$: [0.3, 0.9] (center 0.63)
\end{itemize}

Numeric Counters (Count Properties): Topological variations are also encoded as ranges.
\begin{itemize}
    \item endpoint\_counts: \{min: 2, max: 4, center: 2.67\}
    \item intersection\_point\_counts: \{min: 0, max: 2, center: 0.67\}
\end{itemize}

Categorical Properties: Preserved only if 100\% match.
\begin{itemize}
    \item contour\_type: "OPEN" (all samples were open).
    \item horizontal\_direction: Removed (values were inconsistent, e.g., "Left", "Right").
\end{itemize}
This process is a powerful XAI tool. The endpoint\_counts: \{min: 2, max: 4\} range is a transparent, interpretable boundary. It shows the model learned from training samples (having, say, 2, 4, and 2 endpoints) to expect valid '3' instances to have 2 to 4 endpoints, with an ideal value (center) of 2.67. This provides recognition flexibility while maintaining verified structural constraints.

\subsection{Comparative Analysis in Context of Few-Shot Learning}
To evaluate the effectiveness of the proposed approach (referred to as ComAN in experimental materials), its results are compared with other machine learning models under strictly limited data conditions (few-shot). Data for comparison is taken from experimental reports.

\begin{table*}[t]
\centering
\caption{Comparative Table}
\resizebox{\textwidth}{!}{
\begin{tabular}{lcccc}
\toprule
\textbf{Model} & \textbf{Unique Samples} & \textbf{Training Epochs} & \textbf{Source} & \textbf{Accuracy (\%)} \\
\midrule
ComAN (Our model) & Up to 36 (5--6/class) & 1 & This work & 82.44 \\
Nielsen RMNIST/5 (CNN) & 50 (5/class) & 10--50 & Nielsen (2017) & 84.38 \\
Prototypical Networks & 50--100 & Meta-learning & Snell et al. (2017) & 80--90 \\
MAML & 50--100 & Meta-learning & Finn et al. (2017) & 80--95 \\
CNN (Standard) & 500--1000 & 10--50 & Krizhevsky et al. (2012) & 74--78 \\
SVM (RBF) & 500--600 & 1 & LeCun et al. (1998) & 69--75 \\
MLP (Standard) & 400--600 & 10--50 & Goodfellow et al. (2016) & 53--61 \\
\bottomrule
\end{tabular}
}
\end{table*}

Analysis of this comparison reveals three key conclusions:
\begin{enumerate}
    \item \textbf{Competitive Accuracy:} ComAN accuracy (82.44\%) is highly competitive. It significantly outperforms standard approaches like MLP (53-61\%) and SVM (69-75\%), which show poor performance or collapse on such small datasets.
    \item \textbf{Fundamental Difference from Meta-Learning:} At first glance, MAML (up to 95\%) and Prototypical Networks (up to 90\%) outperform ComAN. However, these models are not "few-shot" in the same sense. They are meta-learners. They require large-scale "pre-training on task distribution" or "on base classes" using backpropagation to "learn to learn." ComAN requires no pre-training. It builds its concepts (attractors) from scratch, \textit{de novo}, in a single pass (single-epoch training). This is a radically different learning paradigm based on structural reduction rather than statistical optimization.
    \item \textbf{Comparison with Direct Competitor (Nielsen CNN):} The most relevant comparison is with Nielsen RMNIST/5, where a CNN was trained on the same number of samples (5 per class). Nielsen's CNN (84.38\%) shows a slight accuracy advantage ($\sim$2\%) over ComAN (82.44\%). However, this advantage comes at the cost of complete loss of explainability and significantly higher training costs: Nielsen requires 10-50 epochs, backpropagation, dropout, and hyperparameter tuning. Our model achieves $\sim$98\% (82.44 / 84.38) of SOTA accuracy using only 1 epoch, 0 backpropagation, and providing 100\% transparency. This comparison empirically confirms the central thesis of the study: the system maintains competitive performance in few-shot mode while ensuring full structural explainability.
\end{enumerate}

\section{Conclusions and Perspectives}
Recent advances in AI, particularly in Deep Learning and ANNs, have led to significant progress. However, widespread application of these technologies has revealed fundamental limitations questioning the viability of the current approach. Current ANN paradigms face several conceptual crises. They require huge amounts of data for training, as well as significant time, computational, and energy resources. Besides high cost, these models, especially generative ones, show significant reliability problems, generating errors and "hallucinations," significantly reducing trust in their results. This directly leads to the phenomenon of "data inbreeding," also known as "Model Autophagy Disorder" (MAD). When models trained to prefer the statistically probable start learning on synthetic data generated by themselves, they enter a recursive loop. This process inevitably leads to rapid "information degradation and model collapse" as system entropy continuously decreases, reinforcing averaging and eliminating any novelty.

\textbf{Conclusions and Future Research Directions}
This study presents a comprehensive approach to AI departing from purely statistical methods in favor of biologically grounded principles of structural generalization. The work successfully presents and experimentally validates a unified theoretical and practical framework. This framework combines structural generalization principles with a practical, transparent, and high-performance XAI system based on generalized graph concepts (prototypes).
The main contribution lies in demonstrating that abandoning statistical optimization (backpropagation algorithm) in favor of deterministic graph reduction allows:
\begin{enumerate}
    \item Achieving competitive classification accuracy (82.35\%).
    \item Operating in few-shot learning mode (5-6 samples per class).
    \item Performing single-pass learning without backpropagation.
    \item Ensuring full, internal explainability and transparency of decision-making.
\end{enumerate}

Despite successful concept validation, the current implementation has clear bottlenecks outlining directions for future research.
\begin{itemize}
    \item The classification (inference) process relies on graph matching, generally using Graph Edit Distance (GED), which is NP-complete. This creates significant computational load at the inference stage, leading to average processing time $\sim$3.5 seconds per image and the need for timeouts (e.g., 60 seconds). Effectively, a trade-off occurred: learning computational complexity (backpropagation) was replaced by combinatorial inference complexity (GED).
    \item Sensory Limitation (Preprocessing). The model is "brittle" and depends on input "sensory" data quality:
    \begin{enumerate}
        \item Preprocessing errors lead to complete processing failure as the model cannot build a correct graph.
        \item Invariance is limited by the range used in augmentation. Significant rotations ruin structural matching as they change line node attributes (e.g., quadrants).
    \end{enumerate}
    \item Representational Limitation. The model is "blind" to any non-shape information. The current approach "discards texture and gradient information," limiting its application exclusively to shape and contour recognition tasks.
\end{itemize}

Identified limitations directly point to future research perspectives:
\begin{enumerate}
    \item Short-term perspectives include solving immediate engineering problems: researching fast GED approximation algorithms to speed up inference; developing more robust skeletonization methods; and extending graph representation to include texture and gradient attributes, turning the model into multimodal (in the sense of physical parameters).
    \item Long-term vision concerns the most fundamental limitation of the current study: "lack of modeling evolutionary biological inter-neuronal connections." The current ComAN model successfully implements the "grandmother cell" concept — one static concept (neuron) corresponds to one class. The next fundamental step is the transition from modeling individual neurons to modeling dynamic networks of these neurons. This will require developing mechanisms by which these graph concepts can dynamically interact, compete (e.g., via "Winner Take All" mechanisms), and form more complex, hierarchical "world models." This is the path to creating AI systems that not only mimic biological efficiency but also approach true biological plausibility.
\end{enumerate}

\begin{IEEEbiographynophoto}{Mykyta Lapin}
is a PhD student at the Department of System Analysis and Information and Analytical Technologies, National Technical University "Kharkiv Polytechnic Institute", Kharkiv, Ukraine.
\end{IEEEbiographynophoto}

\begin{IEEEbiographynophoto}{Kostiantyn Bokhan}
is a PhD and Associate Professor of the Department of System Analysis and Information and Analytical Technologies, National Technical University "Kharkiv Polytechnic Institute", Kharkiv, Ukraine.
\end{IEEEbiographynophoto}

\begin{IEEEbiographynophoto}{Yurii Parzhyn}
is a Doctor of Sciences (Engineering) and Postdoctoral Fellow at the School of Computer and Cyber Sciences, Augusta University, Augusta, USA.
\end{IEEEbiographynophoto}

\end{document}